\documentclass{article}

\usepackage{arxiv}

\usepackage[utf8]{inputenc} 
\usepackage[T1]{fontenc}    
\usepackage{hyperref}       
\usepackage{url}            
\usepackage{booktabs}       
\usepackage{amsfonts}       
\usepackage{nicefrac}       
\usepackage{microtype}      
\usepackage{lipsum}		

\usepackage[utf8]{inputenc}  
\usepackage{graphicx}
\usepackage{cleveref}

\usepackage[square,numbers]{natbib}
\usepackage{doi}

\title{Fine-tuning StyleGAN2 for Cartoon Face Generation}
\author{ 
    {Jihye Back}\\
	Seoul National University \\
	\texttt{100jihye@snu.ac.kr} \\
}
\hypersetup{
pdftitle={CartoonStylegan: Fine-tuning StyleGAN2 for Cartoon Face Generation},
pdfauthor={Jihye Back},
}

\begin{document}
\maketitle

\begin{abstract}
Recent studies have shown remarkable success in the unsupervised image to image (I2I) translation. However, due to the imbalance in the data, learning joint distribution for various domains is still very challenging. Although existing models can generate realistic target images, it's difficult to maintain the structure of the source image. In addition, training a generative model on large data in multiple domains requires a lot of time and computer resources. To address these limitations, we propose a novel image-to-image translation method that generates images of the target domain by finetuning a stylegan2 pretrained model. The stylegan2 model is suitable for unsupervised I2I translation on unbalanced datasets; it is highly stable, produces realistic images, and even learns properly from limited data when applied with simple fine-tuning techniques. Thus, in this paper, we propose new methods to preserve the structure of the source images and generate realistic images in the target domain. The code and results are available at \url{https://github.com/happy-jihye/Cartoon-StyleGan2}

\end{abstract}

\section{Introduction}
The style-based generative model \cite{karras2019style, karras2020analyzing} is very effective in image generation, and it generates high-resolution images by learning not only global attributes but also stochastic details. Several researchers \cite{choi2018stargan, park2019SPADE, isola2017image, liu2017unsupervised, yi2017dualgan, kim2017learning, zhu2017unpaired, abdal2019image2stylegan} have used this model to perform several tasks such as image editing and image translation, noting the powerfulness of style-based architecture. Furthermore, recent research to apply transfer learning to this model is also active. Because style-based architecture trains sequentially from low-resolution to high-resolution, simple fine-tuning techniques make it easier to transfer models of source domains to target domains. Thus, many unsupervised I2I translations have adopted a transfer learning to reduce resources, memory and time.

In this paper, we fine-tune the stylegan2 model for unsupervised I2I translation. We propose two methods to make source images and target images similar, such as paired images.

1. FreezeSG, which freezes the initial blocks of the style vectors and generator. This is very simple and allows the target image to follow the structure of the source image. 

2. Structure Loss, which make to reduce the distance between the inital blocks of the source generator and the inital blocks of the target generator. The effectiveness of applying Layer Swapping to models trained by this loss is also remarkable.

\section{Related Work}
\paragraph{Generative Adversarial Networks (GANs)} The GAN framework \cite{goodfellow2014generative} has shown impressive results in image generation \cite{karras2020analyzing, karras2017progressive, karras2019style}, image translation \cite{choi2018stargan, park2019SPADE, isola2017image, liu2017unsupervised}, image editing \cite{zhu2020domain, kim2021stylemapgan, shen2020closed}, and face image synthesis \cite{karras2020analyzing, karras2019style}. GAN is a generative model that trains through the adversarial process of generator G and discriminator D: The generator creates a fake image to confuse the discriminator so it cannot recognize whether the image is real or fake. In this paper, we aim to align the source distribution to the target domain using the GAN network.

\paragraph{Image-to-Image Translation} Recent studies have shown remarkable success in the image to image (I2I) translation \cite{choi2018stargan, park2019SPADE, isola2017image, liu2017unsupervised} that aims to learn the mapping between an input image and an output image.  Pix2Pix \cite{isola2017image} is the first framework for I2I translation, which uses cGANs \cite{mirza2014conditional} to enable successful I2I transformations for a variety of tasks when there is paired dataset. Because of the limited paired training data, several methods have been proposed to address unsupervised image-to-image translation tasks. UNIT \cite{liu2017unsupervised} assumes a shared latent space, so that image pairs from different domains can be mapped into one shared latent space. Other unsupervised I2I methods \cite{yi2017dualgan, kim2017learning}, including CycleGAN \cite{zhu2017unpaired}, introduce the cycle consistency loss to make a constraint between the source image and the target image. However, these frameworks must be trained on data from both the source domain and the target domain simultaneously. Training consists of the mapping process in multiple domains, which requires a lot of computational resources and time. If there is a severe imbalance between data of each domain, mode collapse may also happen. To overcome these limitations, we proposed a new unsupervised Image-to-Image translation method that generates images of target domains using the stylegan2 \cite{karras2020analyzing} pretrained model.

\begin{figure}[t]
\centering
   \includegraphics[width=17cm]{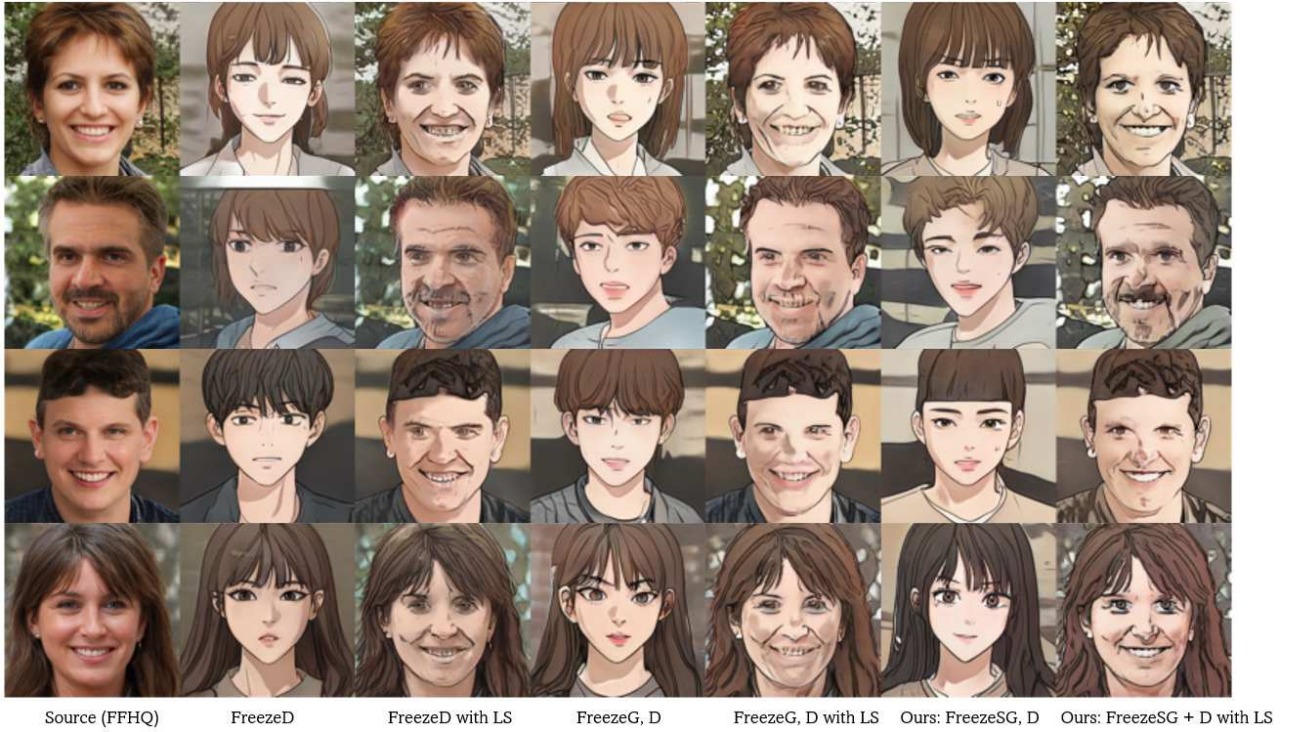}
   \hfil
\caption{We compared the results of FreezeSG with existing models (FreezeD, FreezeG). At this time, we swapped the output of the 4-th style block.}
\label{fig:fig1}
\end{figure} 

\paragraph{Transfer Learning}

Transfer learning is a very powerful and effective method in computer vision. In transfer learning, a new model is trained using the weights of the pre-trained model. This method can reduce time and computer resources because it trains by fine-tuning a pre-trained model without starting from random initialization. Efforts to apply transfer learning to GANs are also active. Mo et al \cite{mo2020freeze} proposed a FreezeD model for the generator to learn the target distribution by freezing the highest-resolution layer of the discriminator during the fine-tuning process. Karras et al \cite{karras2020training} proposed an adaptive discriminator augmentation (ADA) mechanism that stabilizes training in limited data. They have also proven that ADA performs better with FreezeD. These studies have shown that applying transfer learning to GANs allows image translation from the source domain to the target domain more efficiently. 

With the effectiveness and powerfulness of stylegan2 \cite{karras2020analyzing} and fine-tuning techniques \cite{mo2020freeze, karras2020training}, there has been a recent move to apply them to unsupervised I2I translation. FreezeG \cite{lee2020freeze} argued that freezing the generator’s low-level resolution layer helps maintain the structure of the source domain. In addition,  Pinkney et al \cite{pinkney2020resolution} proposed a layer swapping method that combines the high-resolution layer of the FFHQ model with the low-resolution layer of the animation model to generate a photo-realistic face. Huang et al \cite{kwong2021unsupervised} suggested a FreezeFC mechanism that freezes 8-layer $w = f(z), \mathbf{w} \in \mathcal{W} $ of the StyleGAN2 network during the fine-tuning process to increase the similarity between the source and target domains. 

In this paper, we proposed a method for unsupervised image to image(I2I) translation by applying transfer learning to the stylegan2 pretrained model. Inspired by these works \cite{lee2020freeze, pinkney2020resolution, kwong2021unsupervised}, we also proposed new methods to sustain the structure of the source images and generate realistic images in the target domain.

\section{Method}

\subsection{FreezeSG} 
\label{freezesg}

Noting the effectiveness of FreezeG \cite{mo2020freeze}, we analyze the factors that determine the structure of the generated image and find that not only the early layers of the generator but also the style vectors injected into the generator are involved. Inspired by this, we freeze the initial blocks of the generator and the initial style vectors injected in the fine-tuning process of stylegan2. We call this simple yet effective method \emph{a Freeze Style vector and Generator (FreezeSG)}. 

Figure \ref{fig:fig1} shows that FreezeSG reflects the source image better than freezing the generator alone. In addition, it shows that the structure of the source image is better preserved if the low-resolution layer of the generator in the source domain and the high-resolution layer of the generator in the target domain are integrated by applying the Layer Swapping (LS) technique. When LS is applied, the generated images by FreezeSG have a higher similarity to the source image than when FreezeG or the baseline (FreezeD + ADA) were used.

\subsection{Structure Loss}

\begin{figure}[t]
\centering
   \includegraphics[width=17cm]{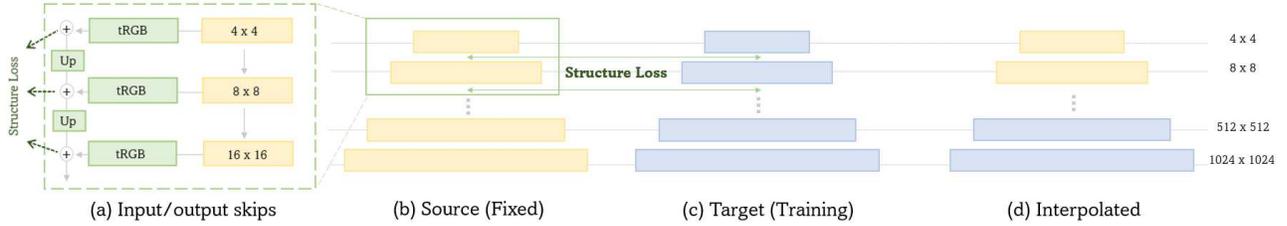}
   \hfil
\caption{The architecture of our generator (using structure loss). If we want to apply a loss for the RGB output of three style blocks(4x4 - 16x16), you calculate the mse-loss between the output of (b) a fixed source generator and the output of (c) the target generator being trained. After calculating, and add them all.(d) shows the integration of the low-resolutoin of the source generator and the high-resolution of the target generator by applying the Layer Swapping method.}
\label{fig:fig2}
\end{figure}

\begin{figure}[t]
\centering
   \includegraphics[width=16cm]{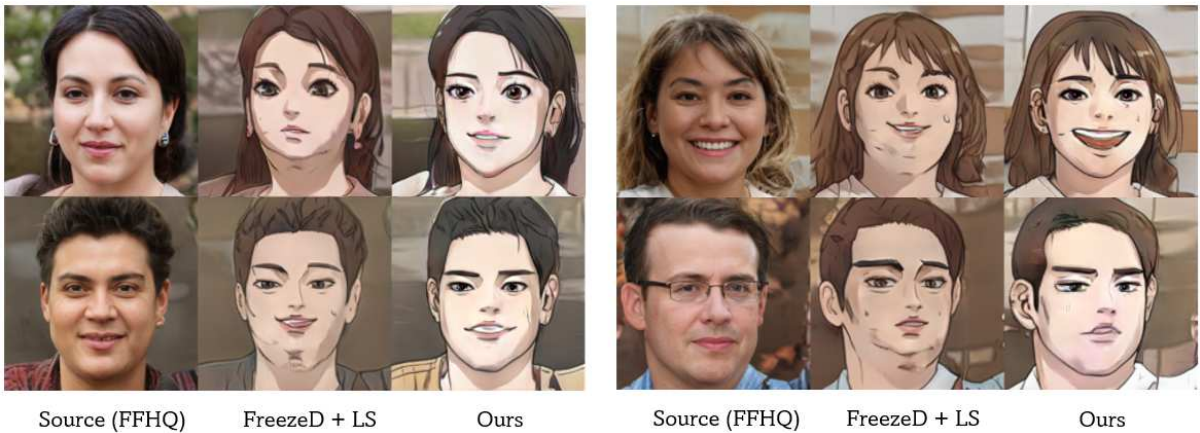}
   \hfil
\caption{Comparison of training results using structure loss and baseline(freezeD). We used the Layer Swapping technique for $l=2$.}
\label{fig:fig3}
\end{figure}

In Section \ref{freezesg}, we showed the effectiveness of FreezeSG. However, since this fixes the weights of the low-resolution layer of the generator, it is difficult to obtain meaningful results when layer swapping on the low-resolution layer. We introduce a simple and effective loss function that we called \emph{structure loss} to obtain meaningful results even when layer swapping is performed for low-resolution layers.

\paragraph{Adversarial Loss.} Original GAN \cite{goodfellow2014generative} is a model  that trains through the adversarial process of generator $G$ and discriminator $D$. The objective of a GAN can be expressed as
\begin{equation}
    \mathcal{L}_{a d v}= \mathbb{E}_{x \sim p_{data}}\left[\log D(x)\right]+\mathbb{E}_{z \sim p(z)}\left[\log \left(1-D(G(z))\right)\right],
\end{equation}
where $G$ tries to minimize this objective to learn the distribution of the target domain, and $D$ tries to maximize it to discriminate between the real image $x$ and the fake image $G(z)$. 

\paragraph{Structure Loss.} 
\label{structureloss}
We adopted the structure of \emph{input/output skips} among the three architectures (MSG-GAN, In-put/output skips, and Residual nets) of the stylegan2 model \cite{karras2020analyzing}. Figure \ref{fig:fig2}(a) shows this architecture, which is simplified by upsampling and summing the contributions of RGB outputs corresponding to different resolutions. Based on the fact that the structure of the image is determined at low resolution, we apply \emph{structure loss} to the values of the low-resolution layer so that the generated image is similar to the image in the source domain. The structure loss makes the RGB output of the source generator to be fine-tuned to have a similar value with the RGB output of the target generator during training.

Structure loss is available as follows.

1. If we want to apply \emph{structure loss} for $n$ style blocks, we must first extract the RGB outputs of both the source generator and the target generator for each resolution.

2. Calculate the mse loss between $G^s_{l=k}(w_s)$ and $G^t_{l=k}(w_s)$ of each resolution, and add it up to the n-th layer.
\begin{equation}
    L_{structure} = \sum_{k=1}^{n}{  \mathbb{E}\left[G^s_{l=k}(w_s)- G^t_{l=k}(w_t)\right]}
\end{equation}
where $G^s_{l=k}(w_s)$ is the RGB outputs of the source generator, and $G^t_{l=k}(w_t)$ is the RGB outputs of the target generator. $k$ is the index of the style block.

\paragraph{Full Objective.} Finally, the objective functions are expressed as
\begin{equation}
 \mathcal{L}_{D}=-\mathcal{L}_{adv}
\end{equation}
\begin{equation}
\mathcal{L}_{G}=\mathcal{L}_{a d v}+\lambda_{structure} \mathcal{L}_{structure}
\end{equation}

where $\lambda_{structure}$ is a hyper-parameter that control the relative importance of structure of source domain. We use $\lambda_{structure}=1$ in all of our experiments.

Figure \ref{fig:fig3} shows that our model generates images much more naturally than baseline \cite{mo2020freeze, karras2020analyzing}. Since we processed the initial layer of $G$ to learn the structure, images of areas such as jaws and heads are well generated.

\begin{figure}[t]
\centering
   \includegraphics[width=17cm]{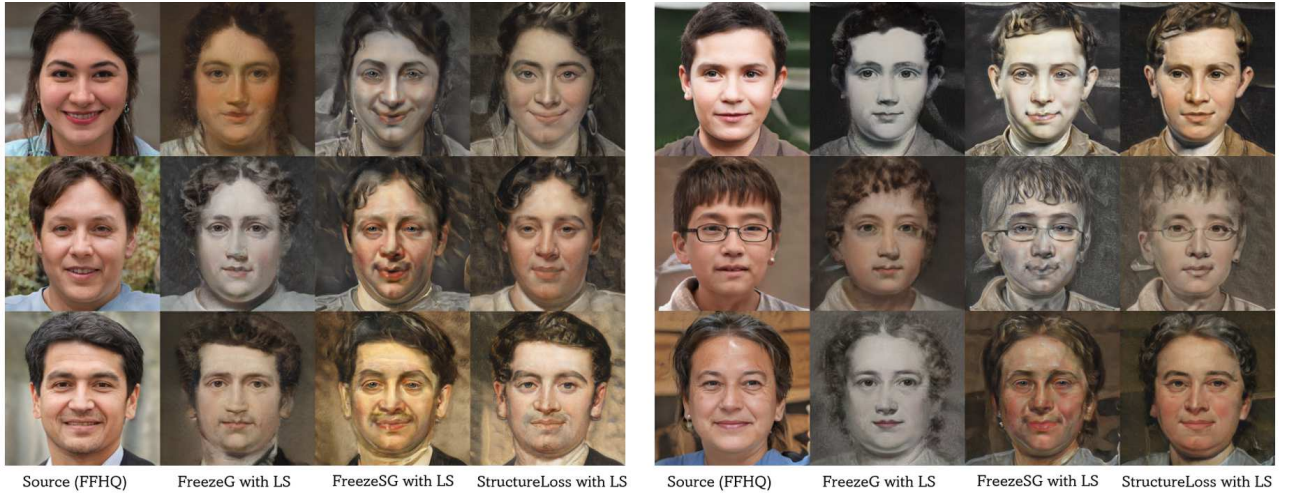}
   \hfil
\caption{Comparison between freezeG and our models.}
\label{fig:fig4}
\end{figure}

\section{Experiments}
\subsection{Datasets}

\paragraph{Source domain dataset.} We used Flickr-Faces-HQ (FFHQ) \cite{karras2019style} as a dataset for the source domain. This has high-quality images of human faces and there are 70,000 images. We used FFHQ to train the stylegan2 model \cite{karras2020analyzing} of 256 resolution.

\paragraph{Target domain dataset.} We experimented with a variety of datasets, including Naver Webtoon \cite{naver}, Metfaces \cite{karras2020training}, and Disney \cite{pinkney2020resolution}. Naver Webtoon Dataset \cite{naver} contains facial images of webtoon characters serialized on Naver. We made this dataset by crawling webtoons from Naver's webtoons site and cropping the faces to 256 x 256 sizes. There are about 15 kinds of webtoons and 8,000 images. We trained the entire Naver Webtoon dataset, and we also trained each webtoon in this experiment. We also experimented with other dataset \cite{karras2020training, pinkney2020resolution}, and learned using images from 256 resolution. 

\subsection{Training details}
In Section \ref{freezesg}, we freeze the initial blocks of the generator and style vector and then train the model by fine-tuning the stylegan2 pre-trained model. In this process, the same objective function as stylegan2 was used. We find it effective to freeze two style blocks (4x4 - 8x8) when generating 256 x 256 resolution images. When applying Layer Swapping \cite{pinkney2020resolution}, it is most effective to integrate the low-resolution layer of the source generator (4x4 - 64x64) and the high-resolution layer (64x64 - 256x256) of the target generator.

\begin{figure}[h]
\centering
   \includegraphics[width=17cm]{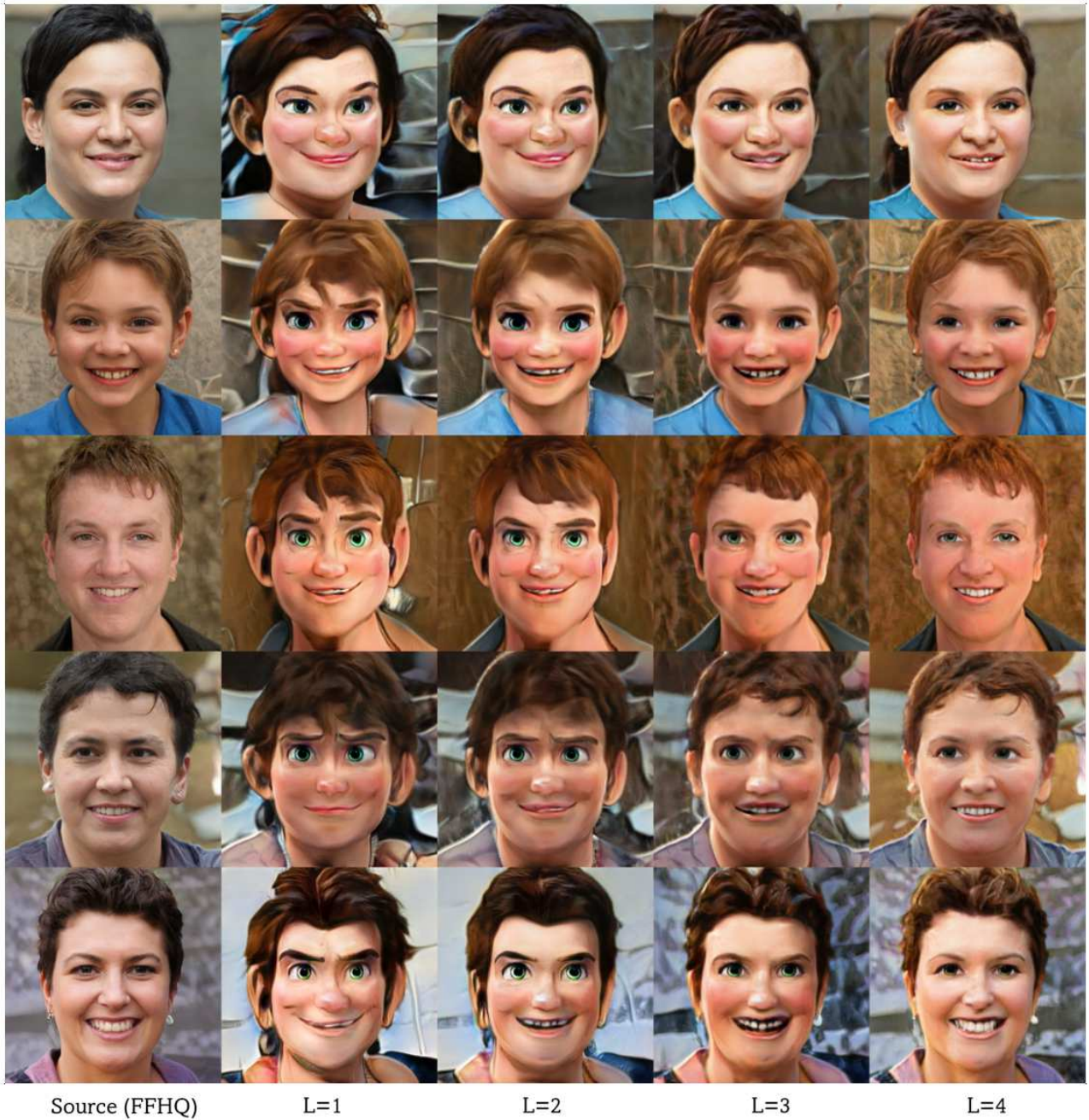}
   \hfil
\caption{The results of applying structure loss with Layer Swapping.}
\label{fig:fig5}
\end{figure}

In Section \ref{structureloss}, We train by applying structure loss to the \emph{input/output skip} architecture of the stylegan2 model \cite{karras2020analyzing}. We used this loss for the three low-resolution layers of the source generator and target generator, and we adopted mse-loss as a loss function. Furthermore, the structural loss enables effective layer swapping even for low-resolution layers, unlike freezeSG. So we swapped different low-resolution layers through various experiments. (Figure \ref{fig:fig5})

\subsection{Experiment Results}
We showed that our model was more effective than the baseline (FreezeD + ADA) \cite{mo2020freeze, karras2020training} and FreezeG \cite{lee2020freeze}. In addition, when the layer swapping technique was applied, the structure of the source domain was more successfully maintained. We also conducted additional experiments on the metface \cite{karras2020training} and disney \cite{pinkney2020resolution} dataset, and the highly quality images were generated when structural loss and layer swapping were used together. FreezeSG also maintained the structure of the source image well but it produced an unnatural image than the image of applying structure loss.

\section{Conclusions and Future works}
In this paper, we proposed Cartoon-StyleGAN: Fine-tuning StyleGAN2 for Cartoon Face Generation, a novel unsupervised Image-to-Image translation methods. Experimental results show that these methods are effective in making source image and target image similar. Compared to previous studies, it generates a more realistic image.

However, our studies need to adjust layers to optimize for each dataset. In future works, we expect to be more stable and performance-enhancing.

\begin{figure}[h]
\centering
   \includegraphics[width=17cm]{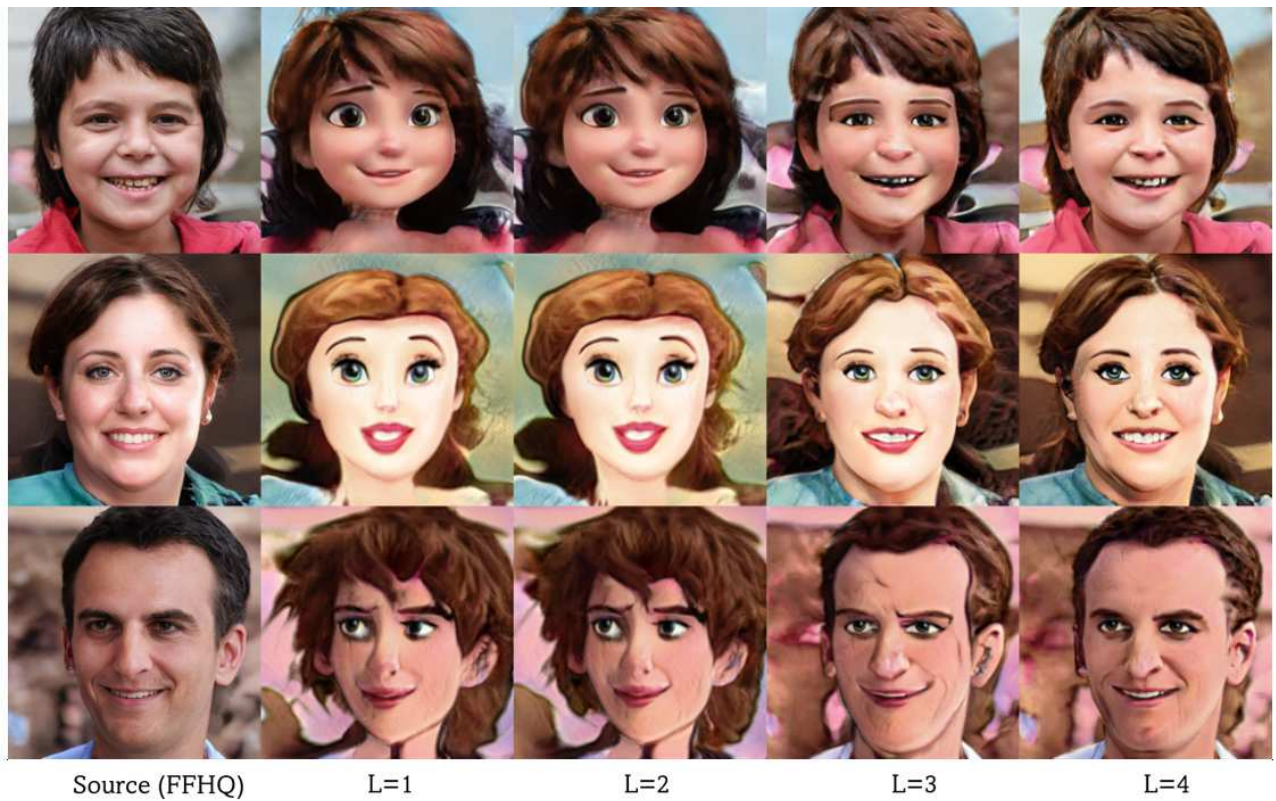}
   \hfil
\caption{The results of FreezeSG with Layer Swapping.}
\label{fig:fig6}
\end{figure}

\bibliographystyle{unsrtnat}
\bibliography{cartoongan.bbl}  

\begin{figure}[t]
\centering
   \includegraphics[width=17cm]{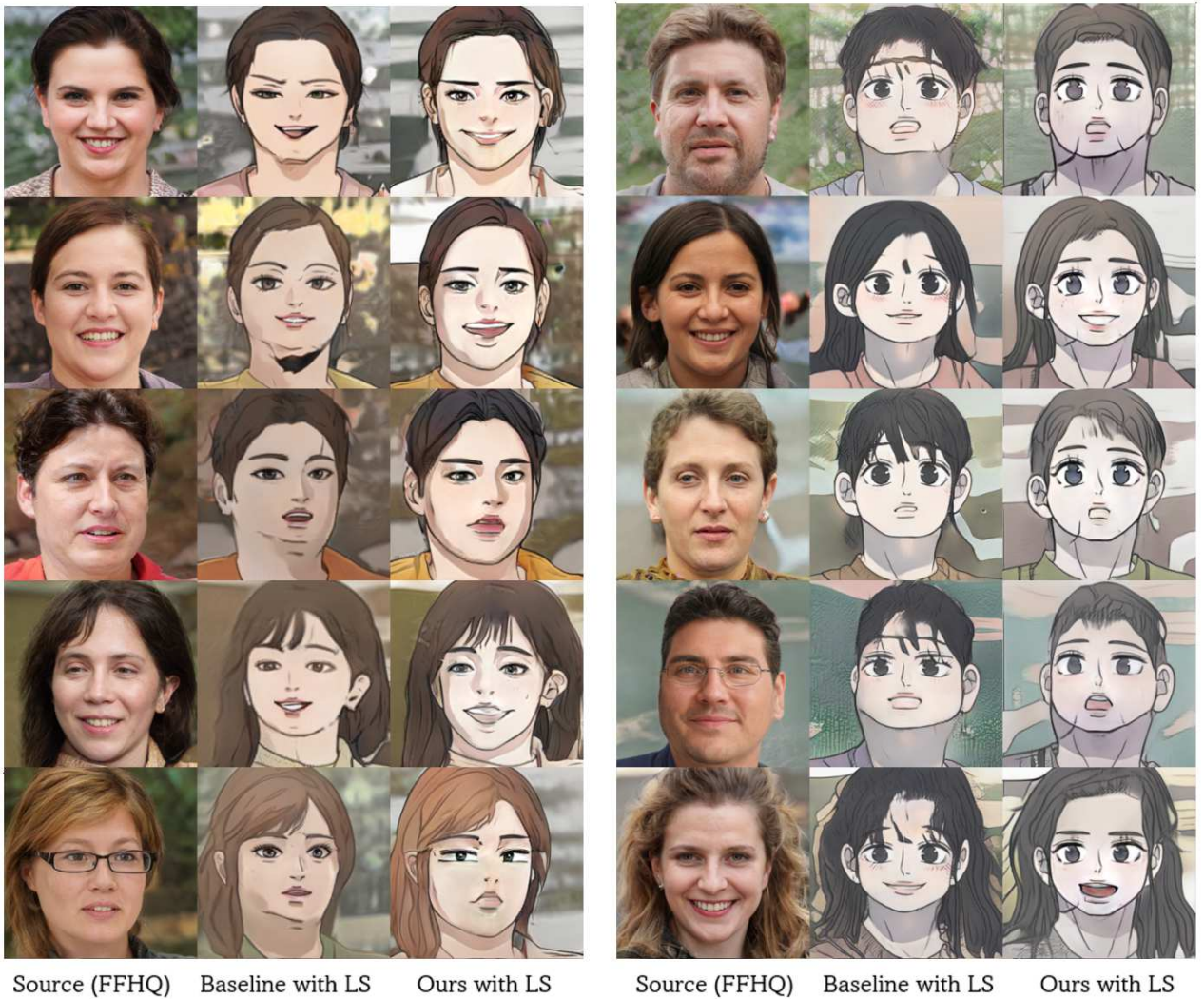}
   \hfil
\caption{ Comparison between baseline(FreezeD + ADA) and our model applying the structure loss.}
\label{fig:fig7}
\end{figure}

\newpage

\appendix

\section{Appendix}
\subsection{Change Facial Expression and Pose.} 

\begin{figure}[h]
\centering
   \includegraphics[width=11cm]{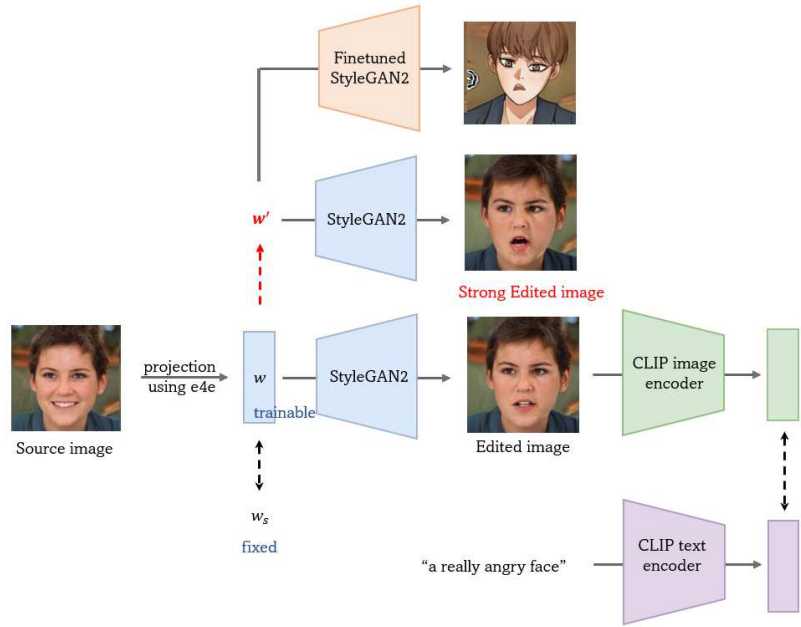}
   \hfil
\caption{The architecture of StyleCLIP.}
\label{fig:fig8}
\end{figure}

\begin{figure}[h]
\centering
   \includegraphics[width=13cm]{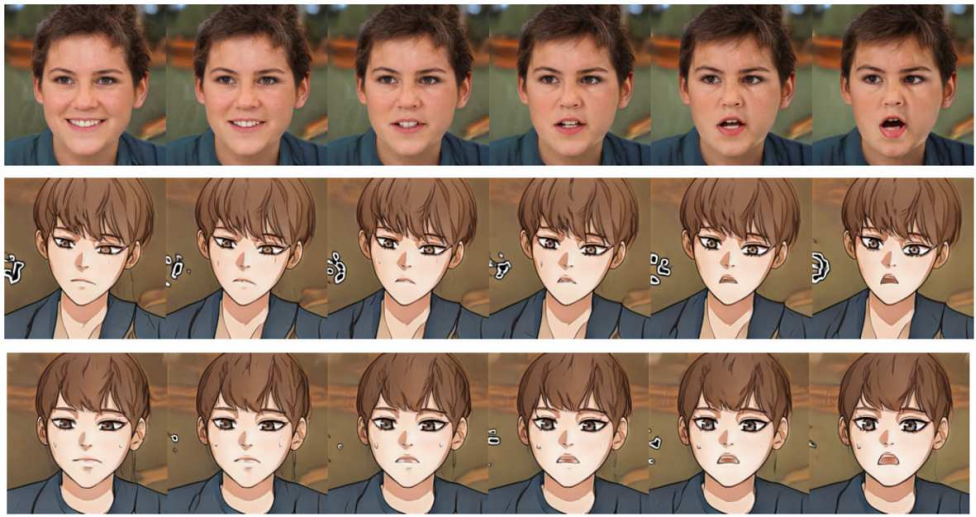}
   \hfil
\caption{The result images of text guided manipulation. We additionally introduce \emph{styleclip strength} to increase image variation. The third image is the one generated by StyleCLIP, and the 6-th image is the one applied with styleclip strength, $\alpha = 2$.}
\label{fig:fig9}
\end{figure}

\paragraph{StyleCLIP}
Inspired by StyleCLIP \cite{patashnik2021styleclip} that manipulates generated images with text, we change the faces of generated cartoon characters by text. We used the latent optimization method among the three methods of StyleCLIP and additionally introduced \emph{styleclip strength}. It allows the latent vector to linearly move in the direction of the optimized latent vector, making the image change better with text.

\begin{equation}
    w' = \alpha (w-w_s) + w)
\end{equation}
where $\alpha$ is a styleclip strength, $w$ is a trainable latent vector, and $w_s$ is a fixed latent vector.

We first found a latent vector optimized according to text using a source generator and styleclip. Then we entered this latent vector into a fine-tuned target generator to change the facial expression of the cartoon face. Figure \ref{fig:fig9} shows that StyleCLIP is effective in changing the expression of a webtoon character.

\end{document}